\documentclass{article}

\PassOptionsToPackage{authoryear}{natbib}

\usepackage[final]{nips_2016}

\usepackage[utf8]{inputenc} % allow utf-8 input
\usepackage[T1]{fontenc}    % use 8-bit T1 fonts
\usepackage{hyperref}       % hyperlinks
\usepackage{url}            % simple URL typesetting
\usepackage{booktabs}       % professional-quality tables
\usepackage{amsfonts}       % blackboard math symbols
\usepackage{nicefrac}       % compact symbols for 1/2, etc.
\usepackage{microtype}      % microtypography
\usepackage{amsmath}
\usepackage{graphicx}
\usepackage{subfig} 
\usepackage{epstopdf}

\title{Hypervolume-based Multi-objective Bayesian Optimization with Student-$t$ Processes}

\author{
  Joachim van der Herten \And Ivo Couckuyt \And Tom Dhaene \\
  IDLab\\
  Ghent University - imec\\
  iGent Tower - Department of Electronics and Information Technology (INTEC)\\
  Technologiepark-Zwijnaarde 15, \\ B-9052 Ghent, Belgium\\
  \texttt{joachim.vanderherten@ugent.be} \\
}
\def\bfx{\mbox{\boldmath$x$}}
\def\bfy{\mbox{\boldmath$y$}}
\def\bfp{\mbox{\boldmath$p$}}
\def\bfu{\mbox{\boldmath$u$}}
\def\bfl{\mbox{\boldmath$l$}}
\def\bff{\mbox{\boldmath$f$}}
\def\bfmu{\mbox{\boldmath$\mu$}}
\def\bftheta{\mbox{\boldmath$\theta$}}
\def\bfphi{\mbox{\boldmath$\phi$}}

\setcitestyle{authoryear}
\begin{document}

\maketitle

\begin{abstract}
  Student-$t$ processes have recently been proposed as an appealing alternative non-parameteric function prior. They feature enhanced flexibility and predictive variance. In this work the use of Student-$t$ processes are  explored for multi-objective Bayesian optimization. In  particular, an analytical expression for the hypervolume-based probability of improvement is developed for independent Student-$t$ process priors of the objectives. Its effectiveness is shown on a multi-objective optimization problem which is known to be difficult with traditional Gaussian processes.
\end{abstract}

\section{Introduction}
The use of Bayesian models and acquisition functions to guide the optimization of expensive, noisy, black-box functions (Bayesian Optimization) has become more popular over the years, and has recently been applied to a wide variety of problems in several fields. The next candidate for evaluation of the computationally expensive black-box function is selected by optimizing an acquisition function relying on Bayesian model(s) approximating the previously observed responses of the black-box function(s). Within Machine Learning, it has for instance been applied to optimize model hyperparameters \citep{Snoek2012,Swersky2014}, as model training involving a lot of data typically makes use of traditional numeric optimization infeasible. 

Within the field of engineering, \citet{Jones1998} introduced the combination of Expected Improvement (EI) and Kriging models for optimization of computer simulations which can take up to several days to perform a single run. This situation is commonly encountered in product design involving, i.a., Computational Fluid Dynamics (CFD) and Finite Element Methods (FEM). Multi-objective (or multi-task) optimization has gained a lot of attention in engineering optimization as product design inherently involves trade-offs as typically several (conflicting) aspects are involved. Frequently used are hypervolume-based acquisition functions such as Hypervolume Expected Improvement \citep{Emmerich2006} or Hypervolume Probability of Improvement (HvPoI) \citep{Couckuyt2014}, assuming a Gaussian process (GP) prior for each objective. More recently, a Multi-objective version of the Predictive Entropy Search has been proposed \citep{Lobato2016}.

Naturally, the correctness of the approximation of the objective(s) is crucial to perform succesful optimization. Erroneous model fits lead to selection of new evaluations based on false beliefs making the discovery of optima unlikely, especially when the input space is large. While GPs have received much attention both as a modeling strategy and within Bayesian optimization, recently Student-$t$ process priors (TP) have been proposed \citep{Hagan1991,Hagan1998,Rasmussen2006,Shah2014} for use with EI. In this contribution we consider the TP prior in the context of multi-objective Bayesian optimization, and develop an analytical expression of the HvPoI acquisition function for it accordingly. TPs have shown to be promising, and their properties such as  additional flexibility and enhanced predictive variance seem to be appealing properties for Bayesian optimization. A brief overview of TPs is given in Section~\ref{sec:stp}, and the formulation of HvPoI assuming TP priors for the objectives is given in Section~\ref{sec:hvpoi}. The performance of this approach is then compared to HvPoI with GP priors in Section~\ref{sec:illustration} on a multi-objective problem.

\section{Student-$t$ Processes}
\label{sec:stp}

Given a $d$-dimensional input space $\mathcal{X} \subset \mathbb{R}^d$, $f$ is a Student-$t$ process with degrees of freedom $\nu > 2$, a continuous mean function $\phi$ and a parametrized kernel function $k$. For any set $X_n \subset \mathcal{X}$ of $n$ inputs $\bfx$, the mapping of these inputs by $f$ is distributed according to a multivariate Student-$t$ distribution: $\bfy = \left( f(\bfx_1), ... f(\bfx_n) \right) \sim \mbox{MVT}_n(\nu, \bfphi, K)$ with $K_{ij} = k (\bfx_i, \bfx_j)$. The likelihood corresponds to the probability density function of an MVT:
\begin{equation}
\label{eq:likelihood}
p(\bfy | X_n, \nu, \bftheta) = \frac{\Gamma \left( \frac{\nu + n}{2} \right)}{\left( \left( \nu - 2 \right) \pi \right) ^ {\frac{n}{2}} \Gamma \left( \frac{\nu}{2}\right)} |K|^{-1/2} \left( 1 + \frac{\beta}{\nu - 2} \right) ^{- \frac{\nu + n}{2}},
\end{equation}
with $\beta=(\bfy-\bfphi)^T K (\bfy-\bfphi)$. \citet{Shah2014} have shown that considering $\bfy|\sigma \sim \mathcal{GP}\left(\phi, (\nu - 2) \sigma\right)$ and marginalizing $\sigma$ out assuming an \textit{inverse Wishart process} prior, recovers Equation~\ref{eq:likelihood}. 

For an arbitrary $\bfx^\star \in \mathcal{X}$ the predictive distribution is also a MVT:
\begin{align}
p(f(\bfx^\star)| X_n, \bfy, \bftheta, \nu) &\sim \mbox{MVT}_1\left(\nu + n, \mu(\bfx^\star), \tilde{s}(\bfx^\star)^2\right), \\
\tilde{s}(\bfx^\star)^2 &= \frac{\nu + \beta - 2}{\nu + n - 2} s(\bfx^\star)^2.
\label{eq:predvar}
\end{align}  
The quantities $\mu$ and $s^2$ are identical to the predictive mean and variance of a GP (assuming the same kernel and parameters). Recent work also shows marginalizing the output scale also yields a related MVT predictive distribution \cite{Gramacy2015,Montagna2016}. This differs from non-analytical marginalization of the kernel lengthscales with Markov chain Monte Carlo methods as applied frequently in Bayesian optimization (see \cite{vanderherten2016} for a comparison of the latter with traditional maximum likelihood estimates).

A fundamental difference is observed in Equation~\ref{eq:predvar}: the variance prediction includes the observed responses, as opposed to GP which only considers the space between inputs. This allows a TP to anticipate changes in covariance structure. Furthermore it was proven that a GP is a special case of a TP, with $\nu \rightarrow \infty$. However, the approach applied for GPs to include noise as part of the likelihood can not be applied for TPs, as the sum of two independent MVTs is not analytically tractable. Instead, a diagonal white noise kernel is added to allow approximation of noisy observations. 
\section{Hypervolume-based Probability of Improvement}
\label{sec:hvpoi}
Given a multi-objective (or multi-task) optimization problem, each evaluated input $\bfx_i$ has $p$ observed reponses $\bfy_i = (f_1(\bfx_i), ..., f_p(\bfx_i))$, together forming a matrix $Y \in \mathbb{R}^{n \times p}$. The rows of this matrix correspond to points in the $p$-dimensional objective space. Of interest are the non-dominated solutions forming the Pareto set $\mathcal{P} \subset Y$. Ideally, we like to find a point $\bfx^\star \in \mathcal{X}: \bfp = f(\bfx^\star)$:
\begin{equation}
\underset{\bfx^\star \in \mathcal{X}}{\max} ~ I(\bfp, \mathcal{P})
\end{equation} 
where $I(.)$ is the improvement function  which is defined in this work using the hypervolume indicator as,
\begin{equation}
I(\bfp, \mathcal{P}) = \begin{cases}
\mathcal{H}(\mathcal{P} \cup \bfp) -\mathcal{H}(\mathcal{P}) & \bfp \in A \\
0 & \mbox{otherwise} 
\end{cases}
\end{equation}
with $A$ the non-dominated section of the objective space, $\mathcal{H}(.)$ defined as the hypervolume of the section of the objective space dominated by the Pareto front (bounded by a reference point $\bff^{\max}$).

\begin{figure}
\centering
\includegraphics[width=0.5\columnwidth]{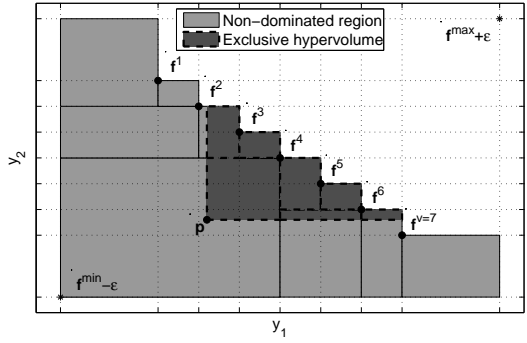}
\caption{Illustration of a Pareto set (members illustrated by $\bff^i$) with two objective functions. $\bff^{\min}$ and $\bff^{\max}$ denote the ideal and anti-ideal point respectively. The shaded areas (both light and dark) represent the non-dominated region and is decomposed into $q$ cells by a binary partitioning procedure. These cells provide integration bounds to compute $I(\bfp, \mathcal{P})$. Courtesy of \citep{Couckuyt2014}}
\label{fig:hvpoi}
\end{figure}
The situation is illustrated in Figure~\ref{fig:hvpoi}: the exclusive (or contributing) hypervolume corresponds to $\mathcal{H}(\mathcal{P} \cup \bfp) -\mathcal{H}(\mathcal{P})$.
Because $\bfp$ is a (black-box) mapping of $p$ objective functions $f$ of an unknown $\bfx^\star$, and because each evaluation is expensive, direct application of traditional numerical optimization methods is infeasible. Instead, we approximate each $f$ and optimize an acquisition function incorporating the information provided by the predictive distributions of the approximations of the objectives. The optimum of the acquisition yields a candidate $\bfx^\star$ to be evaluated on all $f$.

We propose the \textit{Hypervolume Probability of Improvement} as proposed earlier by \citet{Couckuyt2014} as it is tractable and scales to a higher number of objectives, however we assume each $f_i \sim \mathcal{TP}$ instead of a GP. Formally, this acquisition function is defined as 
\begin{align}
P_{\mbox{hv}}(\bfx) &= I(\bfmu, \mathcal{P}) P[\bfx \in A] \\
\bfmu &= [\mu_1(\bfx),...,\mu_p(\bfx)].
\end{align}
 The latter term of the multiplication represents the probability a new point is located in $A$ and, hence, requires an integration over that region. Exact computation of this integral is performed by decomposing $A$ into $q$ cells spanned by upper and lower bounds $[\bfl^k, \bfu^k]$. This decomposition can be done using a binary partition algorithm (which scales poorly as the number of objectives grows) as illustrated in Figure~\ref{fig:hvpoi}, or by applying faster algorithms such as the Walking Fish Group \citep{While2012}. We then make use of the predictive distribution of the TPs:
\begin{equation}
P[\bfx \in A] = \sum_{k=1}^q \prod_{j=1}^p \left( \Phi_{\nu+n}\left( \frac{u^k_j - \mu_j(\bfx)}{\tilde{s}_j(\bfx)} \right) - \Phi_{\nu+n}\left( \frac{l^k_j - \mu_j(\bfx)}{\tilde{s}_j(\bfx)} \right) \right)
\end{equation} 
$\Phi_\nu$ represents the cumulative density function of a $\mbox{MVT}_1(\nu,0,1)$. In addition, we can simply compute the volume of the exclusive volume using the existing $q$ cells with no extra computation as follows (assuming $\bfmu$ is non-dominated):
\begin{equation}
\mathcal{H}(\mathcal{P} \cup \bfmu) - \mathcal{H}(\mathcal{P}) = \sum_{k=1}^q \prod_{j=1}^p \left(  u_j^k - \left( \max l_j^k, \mu_j(\bfx) \right) \right)
\end{equation}

\section{Illustration}
\label{sec:illustration}
We illustrate the effectiveness of the TP prior on the DTLZ1 function, including 6 input parameters and 3 output parameters. The function itself is computed analytically, with some mild Gaussian noise added. \citet{Couckuyt2014} report difficulties approximating the first objective, hence we try the traditional HvPoI in combination with GP priors, and compare it with the modified version as introduced in Section~\ref{sec:hvpoi} with TP priors. The initial set of data points consists of an optimized Latin Hypercube of 10 points. The acquisition function is then permitted to select an additional 30 data points for evaluation. For both TP and GP, the RBF kernel was used, and the hyperparameters $\bftheta$ and $\nu$ were optimized with multi-start Sequential Quadratical Programming. Note that the optimization can result in a very large value $\nu$, causing the TP to become a GP. Hence, we expect better or equal performance, not worse. Both experiments were repeated 10 times.

\begin{figure}[t]
\centering
\subfloat[]{\includegraphics[width=0.43\textwidth]{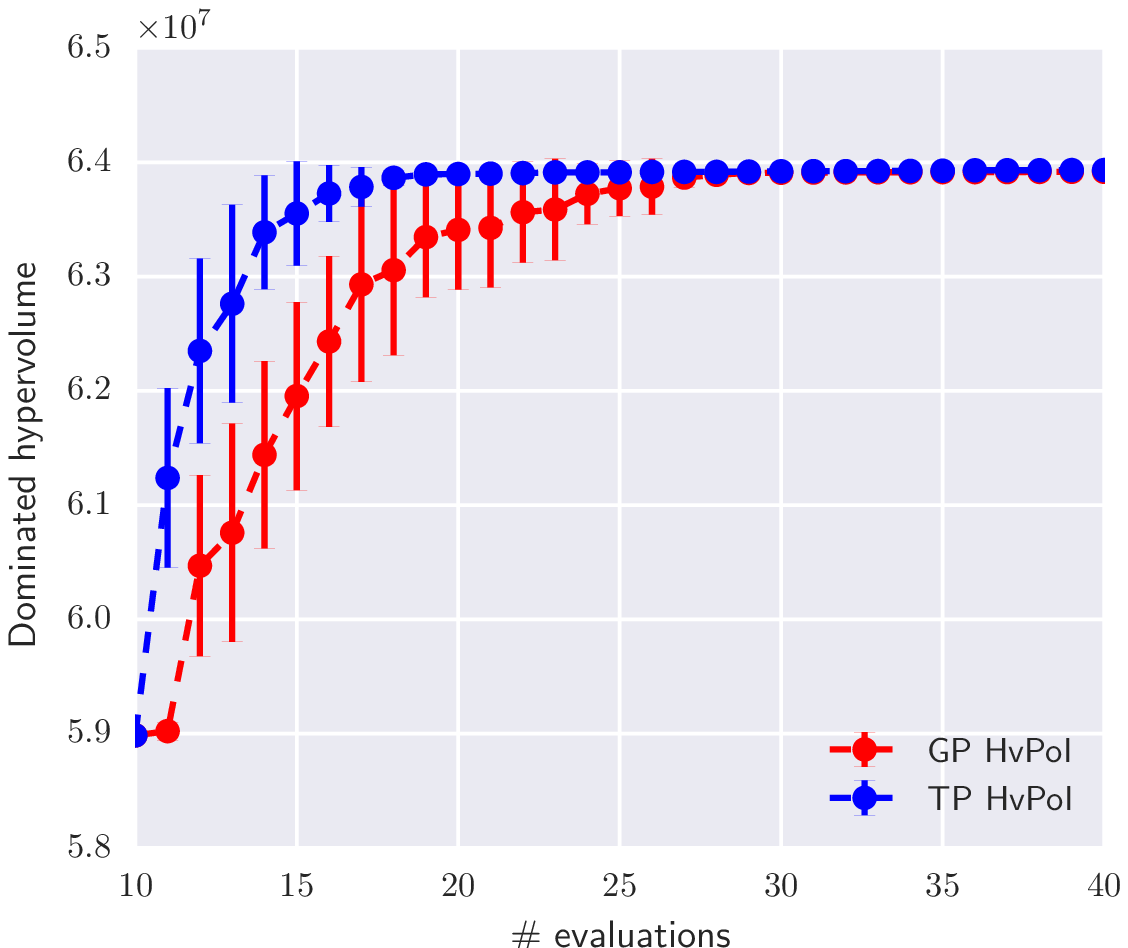}\label{fig:rawresult}}%
~
\subfloat[]{\includegraphics[width=0.43\textwidth]{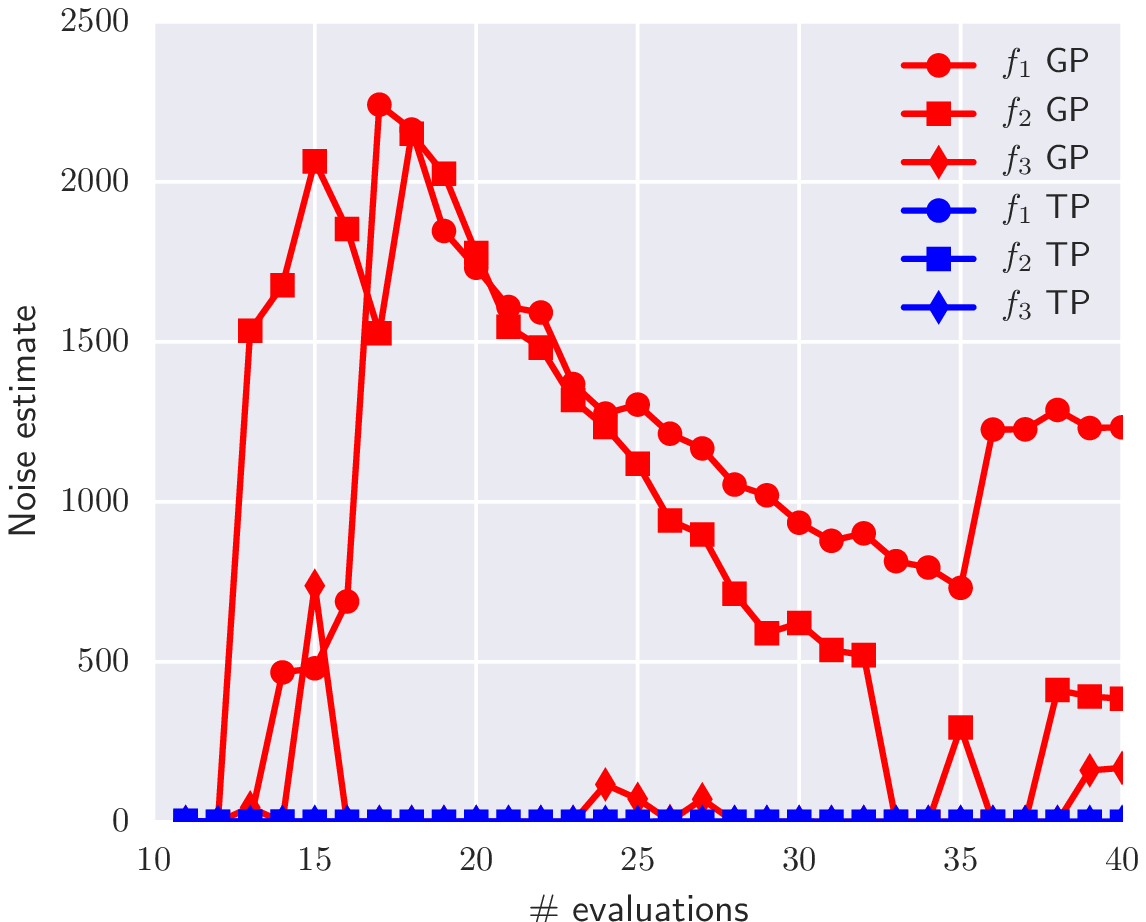}\label{fig:noise}}%
\caption{(a) Comparison of the growth of the dominated hypervolume for the DTLZ1 function, for 10 experiments using both GP and TP priors for the objectives. The mean and 95\% confidence intervals are shown. (b) The noise parameter for all three objectives approximated by GP and TPs. For GPs, the noise is part of the likelihood whereas for TPs a diagonal matrix was added to the kernel matrix. Clearly, the TPs are more flexible and do not consider the evaluated data noisy.}
\label{fig:result}
\end{figure}

As performance metric, the hypervolume indicator (size of the dominated hypervolume with respect to a fixed reference point $\bff^{\max} = [400, 400, 400]$) is recorded after every function evaluation. The average hypervolume and 95\% confidence intervals were computed and plotted in Figure~\ref{fig:rawresult}. Clearly, the runs using the TP approximations of the objectives obtain larger hypervolumes faster. The GP experiments lag behind although they also eventually manage to obtain the same hypervolume indicator performance after additional evaluations. In the end, TPs are able to find a decent hypervolume in about 30\% of the function evaluations needed by the GPs for the same hypervolume indicator performance.

Closer investigation reveals the GP approximations for some of the objective functions have large noise levels, varying significantly as more evaluations are added, whereas the TPs do not as illustrated in Figure~\ref{fig:noise}. It seems the GP is not flexible enough to approximate the objective functions and has to increase the noise variance to avoid ill-conditioning of the kernel matrix. The TPs compensate for this by decreasing the degrees of freedom, which also affects the prediction variance resulting in better selection of evaluation candidates.

\section{Conclusion}
Student-$t$ processes present themselves as an appealing alternative for Gaussian Processes in the context of Bayesian Optimization. Their robustness was proven earlier by \citet{Shah2014} and their enhanced prediction variance can make them more informative for acquisition functions leading to faster discovery of optima. We demonstrated this on a multi-objective optimization problem, using an adapted Hypervolume Probability of Improvement (HvPoI) criterion. 

To make better use of the enhanced prediction variance we aim to adapt the Hypervolume Expected Improvement in further work, as the HvPoI acquisition function does not consider the improvement part of the integration \citep{Couckuyt2014}. In addition other acquisition functions can be modified to be used with TPs, although for some of the more complex acquisition functions the Student-$t$ distribution might introduce tractability challenges. We will be looking at multivariate TPs for multi-objective optimization as in \cite{Lobato2016}. Objectives can then be evaluated indepedently depending on the expected information gain of each.

\subsubsection*{Acknowledgments}
Ivo Couckuyt is a post-doctoral research fellow of the Research Foundation Flanders (FWO).

\bibliographystyle{plainnat}
\bibliography{bibliography}

\begin{thebibliography}{14}
\providecommand{\natexlab}[1]{#1}
\providecommand{\url}[1]{\texttt{#1}}
\expandafter\ifx\csname urlstyle\endcsname\relax
  \providecommand{\doi}[1]{doi: #1}\else
  \providecommand{\doi}{doi: \begingroup \urlstyle{rm}\Url}\fi

\bibitem[Couckuyt et~al.(2014)Couckuyt, Deschrijver, and Dhaene]{Couckuyt2014}
Ivo Couckuyt, Dirk Deschrijver, and Tom Dhaene.
\newblock {Fast calculation of multiobjective probability of improvement and
  expected improvement criteria for Pareto optimization}.
\newblock \emph{Journal of Global Optimization}, 60\penalty0 (3):\penalty0
  575--594, 2014.
\newblock ISSN 0925-5001.

\bibitem[Emmerich et~al.(2006)Emmerich, Giannakoglou, and
  Naujoks]{Emmerich2006}
Michael~TM Emmerich, Kyriakos~C Giannakoglou, and Boris Naujoks.
\newblock {Single-and multiobjective evolutionary optimization assisted by
  gaussian random field metamodels}.
\newblock \emph{IEEE Transactions on Evolutionary Computation}, 10\penalty0
  (4):\penalty0 421--439, 2006.
\newblock ISSN 1089-778X.

\bibitem[Gramacy and Apley(2015)]{Gramacy2015}
Robert~B Gramacy and Daniel~W Apley.
\newblock {Local Gaussian process approximation for large computer
  experiments}.
\newblock \emph{Journal of Computational and Graphical Statistics}, 24\penalty0
  (2):\penalty0 561--578, 2015.

\bibitem[Hern\'{a}ndez-Lobato et~al.(2016)Hern\'{a}ndez-Lobato,
  Hern\'{a}ndez-Lobato, Shah, and Adams]{Lobato2016}
Daniel Hern\'{a}ndez-Lobato, Jos\'{e}~Miguel Hern\'{a}ndez-Lobato, Amar Shah,
  and Ryan~P Adams.
\newblock {Predictive Entropy Search for Multi-objective Bayesian
  Optimization}.
\newblock In Maria~Florina Balcan and Kilian~Q. Weinberger, editors,
  \emph{{Proceedings of the 33rd International Conference on Machine Learning
  (ICML-16)}}, pages 1492--1501. JMLR Workshop and Conference Proceedings,
  2016.

\bibitem[Jones et~al.(1998)Jones, Schonlau, and Welch]{Jones1998}
D.~R. Jones, M.~Schonlau, and W.~J. Welch.
\newblock {Efficient Global Optimization of Expensive Black-Box Functions}.
\newblock \emph{J. of Global Optimization}, 13\penalty0 (4):\penalty0 455--492,
  1998.
\newblock ISSN 0925-5001.

\bibitem[Montagna and Tokdar(2016)]{Montagna2016}
Silvia Montagna and Surya~T Tokdar.
\newblock {Computer Emulation with Nonstationary Gaussian Processes}.
\newblock \emph{SIAM/ASA Journal on Uncertainty Quantification}, 4\penalty0
  (1):\penalty0 26--47, 2016.

\bibitem[O'Hagan(1991)]{Hagan1991}
Anthony O'Hagan.
\newblock {Bayes--hermite quadrature}.
\newblock \emph{Journal of statistical planning and inference}, 29\penalty0
  (3):\penalty0 245--260, 1991.

\bibitem[O'Hagan et~al.(1998)O'Hagan, Bernardo, Berger, Dawid, Smith,
  et~al.]{Hagan1998}
Anthony O'Hagan, JM~Bernardo, JO~Berger, AP~Dawid, AFM Smith, et~al.
\newblock {Uncertainty analysis and other inference tools for complex computer
  codes}.
\newblock 1998.

\bibitem[Rasmussen(2006)]{Rasmussen2006}
Carl~Edward Rasmussen.
\newblock {Gaussian processes for machine learning}.
\newblock 2006.

\bibitem[Shah et~al.(2014)Shah, Wilson, and Ghahramani]{Shah2014}
Amar Shah, Andrew~Gordon Wilson, and Zoubin Ghahramani.
\newblock {Student-t Processes as Alternatives to Gaussian Processes.}
\newblock In \emph{{AISTATS}}, pages 877--885, 2014.

\bibitem[Snoek et~al.(2012)Snoek, Larochelle, and Adams]{Snoek2012}
Jasper Snoek, Hugo Larochelle, and Ryan~P Adams.
\newblock {Practical Bayesian optimization of machine learning algorithms}.
\newblock In \emph{{Advances in neural information processing systems}}, pages
  2951--2959, 2012.

\bibitem[Swersky et~al.(2014)Swersky, Snoek, and Adams]{Swersky2014}
Kevin Swersky, Jasper Snoek, and Ryan~Prescott Adams.
\newblock {Freeze-thaw Bayesian optimization}.
\newblock \emph{arXiv preprint arXiv:1406.3896}, 2014.

\bibitem[van~der Herten et~al.(2016)van~der Herten, Couckuyt, Deschrijver, and
  Dhaene]{vanderherten2016}
Joachim van~der Herten, Ivo Couckuyt, Dirk Deschrijver, and Tom Dhaene.
\newblock {Fast Calculation of the Knowledge Gradient for Optimization of
  Deterministic Engineering Simulations}.
\newblock \emph{arXiv preprint arXiv:1608.04550}, 2016.

\bibitem[While et~al.(2012)While, Bradstreet, and Barone]{While2012}
Lyndon While, Lucas Bradstreet, and Luigi Barone.
\newblock {A fast way of calculating exact hypervolumes}.
\newblock \emph{IEEE Transactions on Evolutionary Computation}, 16\penalty0
  (1):\penalty0 86--95, 2012.
\newblock ISSN 1089-778X.

\end{thebibliography}

\end{document}